\documentclass[lettersize,journal]{IEEEtran}
\usepackage{amsmath,amsfonts}
\usepackage{algorithm}
\usepackage{algpseudocode} 
\usepackage{amssymb} 
\usepackage{array}
\usepackage[caption=false,font=normalsize,labelfont=sf,textfont=sf]{subfig}
\usepackage{textcomp}
\usepackage{stfloats}
\usepackage{url}
\usepackage{verbatim}
\usepackage{cite}
\usepackage{booktabs}   
\usepackage{graphicx} 
\begin{document}

\title{QAROO: AI-Driven Online Task Offloading for Energy-Efficient and Sustainable MEC Networks}

\author{
    Yongtao Yao,
    Yao Yang,
    Haorui Shi,
    Canglu Zhu,
    Miaojiang Chen,
    Ahmed Farouk
\thanks{Corresponding author: Miaojiang Chen (email: mjchen\_cs@gxu.edu.cn).}

\thanks{Manuscript received Date Month Year; revised Date Month Year.}
}

\markboth{IEEE Transactions on ...}%
{Yao \MakeLowercase{\textit{\textit{et al.}}}: QAROO: AI-Driven Online Task Offloading for MEC}


\maketitle

\begin{abstract}
With the rapid advancement of artificial intelligence (AI) and intelligent science, intelligent edge computing has been widely adopted. However, the limitations of traditional methods, such as poor adaptability and the slow convergence of heuristic algorithms, are becoming increasingly evident. To enable sustainable and resource-efficient edge applications, this paper proposes an online task offloading framework for wireless powered mobile edge computing (MEC) networks, called Quantum Attention-based Reinforcement learning for Online Offloading (QAROO). The system employs a binary offloading strategy with the aim of co-optimizing computing and energy resources in dynamic channel environments. In response to the issues of poor adaptability in traditional approaches and the slow convergence of heuristic algorithms, the framework integrates quantum neural networks and attention mechanisms, introducing three key improvements: using recurrent neural networks to enhance temporal modeling capability, proposing an uncertainty-guided quantization method to improve exploration efficiency, and incorporating attention mechanisms into quantum networks to strengthen feature representation. Experiments demonstrate that the proposed method outperforms comparative schemes in terms of normalized computation speed and processing time, offering an efficient and stable solution for online task offloading in large-scale Internet of Things (IoT) dynamic environments.
\end{abstract}

\begin{IEEEkeywords}
Quantum Attention Mechanism, Task Offloading, Recurrent Neural Network, Wireless Powered, Sustainable Computing.
\end{IEEEkeywords}

\section{Introduction}\label{sec1}
Size-constrained devices face bottlenecks in battery life and computational capability. Wireless power transfer (WPT) enables sustained radio frequency (RF) energy charging over the air, while  MEC technology allows devices to offload computation-intensive tasks to nearby edge servers (ES), significantly reducing computational latency and local energy consumption \cite{7762913}. The integration of these two technologies, known as wireless-powered mobile edge computing, can address these bottlenecks. In such systems, an access point (AP) simultaneously transmits RF energy to devices to sustain their operation and receives their offloaded computational tasks for remote processing. Devices typically employ a binary offloading policy \cite{8016573}, meaning a task is either computed locally or fully offloaded to the MEC server.

In multi-user MEC networks, due to the presence of binary offloading variables, opportunistic computation offloading optimization requires the formulation of a mixed-integer programming (MIP) model to jointly determine the binary offloading decisions and the allocation of resources for wireless energy transfer and task offloading time. However, solving such problems is extremely complex, especially in large-scale networks. Numerous studies have focused on designing low-complexity suboptimal algorithms, such as heuristic algorithms based on local search \cite{8016573}, decomposition-oriented search algorithms \cite{8334188}, and convex relaxation methods for binary variables \cite{8352664}, among others. Nonetheless, besides performance loss, these algorithms often require extensive numerical iterations to obtain satisfactory solutions. Consequently, they are unsuitable for real-time decision-making in fast-fading channel environments, where the offloading strategy needs to be constantly updated in response to significant changes in the channel state.

\begin{figure}[htbp]
    \centering
    \includegraphics[width=0.5\textwidth]{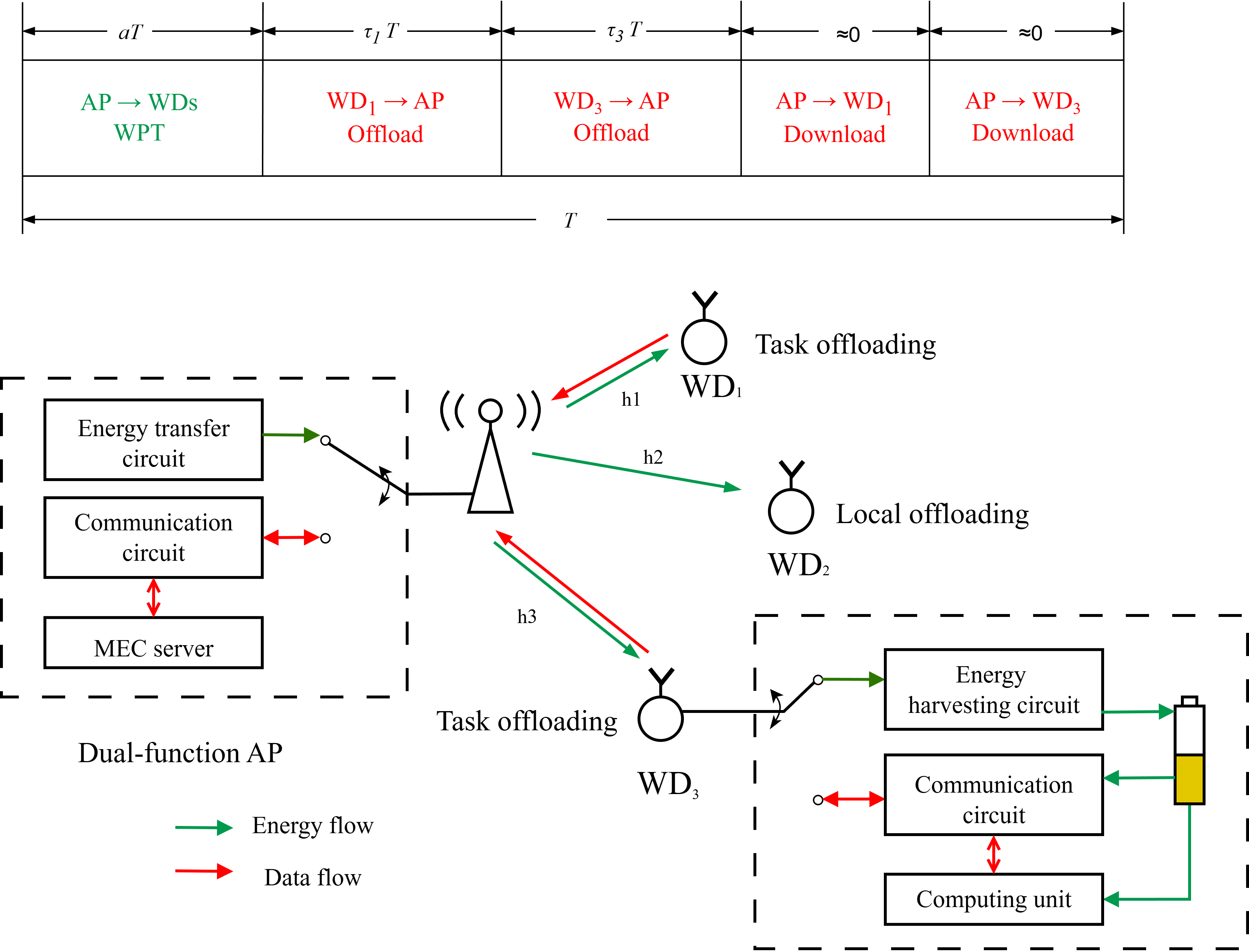} 
    \caption{An example of the considered wireless powered MEC network
and system time allocation.} 
    \label{fig:fi2_model}
\end{figure}

As shown in \ref{fig:fi2_model}, this paper considers a wireless-powered MEC network comprising one AP and multiple wireless devices, where each wireless device adopts a binary offloading policy. Our objective is to jointly optimize, according to time-varying wireless channels, the task offloading decisions for individual wireless devices, the transmission time allocation between wireless energy transfer and task offloading, and the time allocation among multiple wireless devices. To this end, we propose a QAROO framework to maximize the weighted sum computation rate of all wireless devices. The innovative contributions of this paper are as follows:
\begin{itemize}
\item We innovatively propose the QAROO framework. This framework can autonomously optimize its action generation policy by learning from historical offloading experiences under different wireless fading channel conditions. Compared to the traditional (Deep Reinforcement Learning for Online
Computation Offloading in Wireless Powered
Mobile-Edge Computing Networks)DROO framework, QAROO achieves a significant improvement in computational rate, effectively reduces computational loss, and demonstrates superior and more stable performance in both small-scale and large-scale user scenarios.
\item Building upon the classical DROO framework, we replace the Deep Neural Network (DNN) with a Recurrent Neural Network (RNN). Experimental results show that the improved model converges faster in terms of normalized rate, and the loss function value after convergence is only half that of the original framework.
\item Regarding the order-preserving quantization method, we propose a novel diversity-preserving quantization strategy. This method generates diverse candidate actions through dynamic threshold adjustment and random perturbation, and proactively flips dimensions with higher uncertainty upon detecting action repetition to maintain exploration diversity. Compared to traditional order-preserving quantization methods, this uncertainty-guided mechanism significantly enhances the diversity of the action space and the overall convergence performance of the algorithm.
\end{itemize}
The remainder of this paper is organized as follows.Section \ref{sec2} reviews the related work on quantum neural networks, attention mechanisms, and their integration in reinforcement learning for task offloading.Section \ref{sec3} presents the system model of the wireless-powered MEC network and formulates the mixed-integer optimization problem.Section \ref{sec4} details the proposed QAROO framework, including the RNN-based temporal modeling, uncertainty-guided quantization, and the quantum‑attention hybrid architecture.Section \ref{sec5} evaluates the performance of the proposed method through simulation experiments and compares it with baseline algorithms.Finally, Section \ref{sec6} concludes the paper and outlines future research directions.

\section{Related Work}\label{sec2}

Quantum Neural Network (QNN) has emerged as a promising paradigm for leveraging quantum computing to enhance machine learning tasks. These networks are designed to operate on quantum data and leverage quantum principles such as superposition and entanglement to perform computations that are challenging for classical neural networks. Beer et al. \cite{WOS:000514356400004} propose a training methodology for deep quantum neural networks, while Gao and Duan \cite{WOS:000411526700007} develop efficient neural network representations for quantum many-body states. Gong et al. \cite{WOS:001158795100001} present a quantum convolutional neural network (QCNN) based on variational quantum circuits, achieving faster convergence and higher accuracy in image classification. Cong et al. \cite{WOS:000500574300022} further design QCNNs that demonstrate advantage in quantum phase recognition tasks. Abbas et al. \cite{WOS:000888558600009} analyze the power of quantum neural networks, showing that they may possess superior learning capabilities on near-term quantum hardware. Deng et al. \cite{WOS:000401235900001} integrate entanglement analysis with neural network states, underscoring the potential of QNNs in modeling highly entangled quantum systems.

Attention Mechanisms have become a cornerstone of modern deep learning, enabling models to focus on the most relevant parts of input data. Inspired by human cognitive processes, attention mechanisms enhance the performance of neural networks across various domains, including natural language processing, computer vision, and reinforcement learning. Niu et al. \cite{WOS:000663092000005} and Soydaner \cite{WOS:000801057500001} provide comprehensive reviews of attention mechanisms, while Correia and Colombini \cite{WOS:000763844800001} survey neural attention models across deep learning architectures. In natural language processing, Galassi et al. \cite{WOS:000704111000006} show that attention models are particularly effective in capturing long-range dependencies and improving interpretability. The flexibility and power of attention mechanisms make them a vital component in advancing the capabilities of classical and, more recently, quantum machine learning models.

Integration of Quantum Neural Network with Attention Mechanisms represents a cutting-edge research direction aimed at harnessing the strengths of both quantum computing and attention-based learning. Zhao et al. \cite{WOS:001364431200167} propose the Quantum Kernel Self-Attention Network (QKSAN), which combines quantum kernel methods with self-attention to enhance data representation. Shi et al. \cite{WOS:001381551700001} present the Quantum Self-Attention Network (QSAN), a near-term achievable architecture for quantum attention. Li and Ruan \cite{WOS:001414377800006} apply QSAN to combinatorial optimization, demonstrating its effectiveness in the traveling salesman problem. Hsu et al. \cite{11249855} introduce the Quantum Adaptive Excitation Network (QAE-Net), which integrates variational quantum circuits into classical attention modules to capture higher-order dependencies. These advances indicate that quantum-enhanced attention mechanisms not only boost learning efficiency but also pave the way for scalable quantum machine learning applications in the NISQ era.

Recurrent Neural Network (RNNs) are designed to process sequential data by retaining a hidden state that captures temporal dependencies, making them effective for time-series and sequential decision tasks \cite{ELMAN1990179}. Advanced variants such as Long Short-Term Memory (LSTM) \cite{6795963} and Gated Recurrent Unit (GRU) address the vanishing gradient problem and enhance long-term memory. In reinforcement learning and dynamic resource allocation, RNNs help model partially observable environments and leverage historical states for decision-making \cite{hausknecht2015deep}. Their ability to learn from past states is particularly advantageous in non-stationary wireless channels, where channel conditions and offloading histories evolve. Replacing standard DNNs with RNNs has been shown to improve convergence and stability in memory-sensitive scenarios \cite{SHERSTINSKY2020132306}. This motivates our use of an RNN in the DROO framework \cite{8771176} to better capture temporal patterns in wireless environments and enhance online offloading performance.

\section{System Model}\label{sec3}
To ensure a fair and consistent performance comparison with the original DROO \cite{8771176}framework, this paper fully adopts the wireless-powered mobile edge computing system model proposed therein. The model describes a static network scenario comprising a single AP and multiple wireless devices, with the core objective of maximizing the system-wide weighted sum computation rate under time-varying wireless channel conditions through the joint optimization of binary task offloading decisions and continuous wireless resource allocation. Consider a network as shown in figure \ref{fig:fi2_model}, which includes an AP equipped with a stable power supply and a set of $N$ WDs. Each WD is equipped with a single antenna and a rechargeable battery, capable of harvesting energy from the RF signals broadcast by the AP. The AP has a dual function: it acts as a power station for wireless energy transfer to downstream WDs and also as an edge server for receiving and processing the computation tasks offloaded by the WDs. This architecture is suitable for static or low-mobility IoT scenarios such as wireless sensor networks. The system operates in the same frequency band and employs time-division duplexing to avoid interference between energy transfer and data communication.

The system operates on a discrete timeline, where time is divided into equal-length frames of duration $T$. The frame length $T$ is smaller than the channel coherence time, so it is reasonable to assume that the wireless channel gain between the AP and each WD remains constant within each time frame but varies independently across different frames. Let $h_i$ denote the channel power gain of the $i$-th WD in the current frame, determined jointly by the average path loss and small-scale fading following a Rayleigh distribution. The operation within each time frame is divided into two stages. At the beginning of the frame, the AP performs wireless energy transfer at a fixed power $P$ for a duration of $aT$, where $a \in [0,1]$ is the energy transfer time ratio to be optimized. The energy harvested by the $i$-th WD during this period is 

\begin{equation}
E_i = \mu P h_i a T,
\end{equation}

Where $\mu$ is the energy harvesting efficiency. After energy harvesting, each WD needs to process an indivisible computation task. The system adopts a binary offloading policy, meaning each task is either entirely executed locally or entirely offloaded to the AP. This decision is represented by a binary variable $x_i \in \{0,1\}$. Each WD is assigned a weight $w_i$ to reflect its priority in the system objective.

If a WD chooses local computation ($x_i=0$), it utilizes the harvested energy to operate continuously throughout the frame. Under the constraint that the processor energy consumption follows the $k_i f_i^3$ model, by optimizing the computing frequency, its maximum local computation rate $r_{L, i}^*(a)$ can be derived as

\begin{equation}
r_{L,i}^*(a) = \eta_1 \left( \frac{h_i}{k_i} \right)^{\frac{1}{3}} a^{\frac{1}{3}},
\end{equation}

where $\eta_1 \triangleq (\mu P)^{\frac{1}{3}} / \phi$, which is a fixed parameter.

If a WD chooses offloading computation ($x_i=1$), it uses all the harvested energy to transmit the task data to the AP. Let $\tau_i$ denote the allocated task transmission time ratio for this WD; then its computation rate equals the data transmission rate $r_{O,i}^*(a, \tau_i)$, given by

\begin{equation}
r_{O,i}^*(a, \tau_i) = \frac{B \tau_i}{v_u} \log_2 \left( 1 + \frac{\mu P h_i^2 a}{\tau_i N_0} \right),
\end{equation}

where $B$ is the bandwidth, $N_0$ is the noise power, and $v_u$ is a constant. Since the AP is assumed to have a strong computational capability and the result data size is small, its processing time and the downlink transmission time are negligible. Therefore, the time in each frame is shared only between the energy transfer time and the task transmission times of all WDs, satisfying the constraint.

\begin{equation}
a + \sum_{i=1}^{N} \tau_i \leq 1.
\end{equation}

Based on the above model, in each time frame, given the current channel state vector $\mathbf{h} = [h_1, \ldots, h_N]$, the system optimization problem is formulated as a mixed-integer programming problem (P1):

\begin{align}
(P1): \quad & \max_{\mathbf{x}, \boldsymbol{\tau}, a} \; Q(\mathbf{h}, \mathbf{x}, \boldsymbol{\tau}, a) = \sum_{i=1}^{N} w_i \left[ (1-x_i) r_{L,i}^*(a) + x_i r_{O,i}^*(a, \tau_i) \right] \\
\text{s.t.} \quad & a + \sum_{i=1}^{N} \tau_i \leq 1, \nonumber \\
& a \geq 0, \; \tau_i \geq 0, \; \forall i, \nonumber \\
& x_i \in \{0,1\}, \; \forall i. \nonumber
\end{align}

The decision variables include the binary offloading vector \(\mathbf{x}\), the transmission time vector \(\boldsymbol{\tau}\), and the energy transfer time \(a\). The difficulty of this problem lies in the coupling between integer and continuous variables. However, once the offloading decision \(\mathbf{x}\) is fixed, the original problem reduces to a convex optimization subproblem (P2) concerning the continuous variables \(\{a, \boldsymbol{\tau}\}\):

\begin{equation}
\begin{aligned}
\text{(P2):} \quad & \max_{\boldsymbol{\tau}, a} \; Q(\mathbf{h}, \mathbf{x}, \boldsymbol{\tau}, a) = \sum_{i=1}^{N} w_i \left[ (1-x_i) r^{*}_{L,i}(a) + x_i r^{*}_{O,i}(a, \tau_i) \right] \\
\text{s.t.} \quad & a + \sum_{i=1}^{N} \tau_i \leq 1, \\
& a \geq 0, \; \tau_i \geq 0, \; \forall i \in \{1,\ldots,N\}.
\end{aligned}
\end{equation}

This subproblem can be efficiently solved by convex optimization methods (such as the bisection search based on the Lagrangian dual).

Therefore, the key and bottleneck to solving the original problem (P1) lies in how to quickly generate a high-quality (near-optimal) offloading decision \(\mathbf{x}\) for each time-varying channel state \(\mathbf{h}\). All subsequent algorithmic improvements proposed in this paper will focus on this upper-level decision problem, while the lower-level resource allocation subproblem (P2) will serve as an efficient and reliable black-box evaluation function to measure the system performance achievable by any given offloading decision \(\mathbf{x}\).
\section{Proposed Method}\label{sec4}

\begin{figure*}[htbp]
    \centering
    \includegraphics[width=\textwidth]{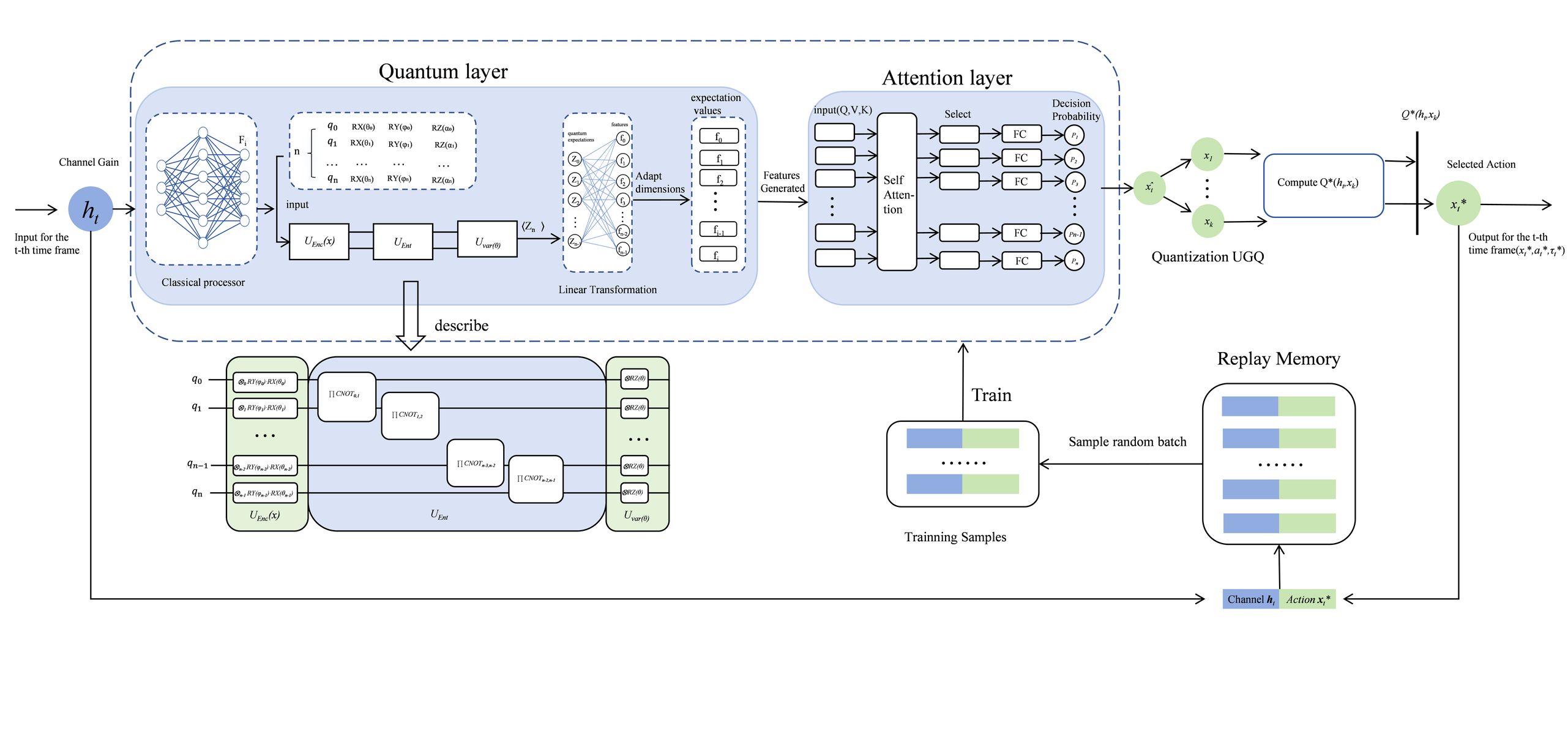}
    \caption{Overall architecture of the proposed QAROO framework.}
    \label{fig:a}
\end{figure*}

Figure \ref{fig:a} illustrates the overall architecture of the proposed QAROO framework. The framework consists of three core components: (1) a RNN for temporal modeling of channel states; (2) a UGQ module for diverse action generation; and (3) a QNN+Attention for enhanced feature representation. The system operates in an online manner where, at each time frame, the current channel state $\mathbf{h}$ is fed into the RNN to generate offloading preferences. The UGQ module then converts these continuous preferences into multiple discrete offloading actions $\{\mathbf{x}_i\}$, which are evaluated by solving the resource allocation subproblem (P2). The best action is selected and executed, while the experience tuple $(\mathbf{h}, \mathbf{x}^*)$ is stored in a replay buffer for subsequent training of the QNN+Attention network. This iterative process enables adaptive decision-making in dynamic wireless environments.
\subsection{Reinforcement Learning Formulation}

Problem (P1) can be reformulated into a Markov Decision Process (MDP). In the MDP, the AP acts as an agent to learn an optimal policy that maximizes the long-term weighted sum computation rate. The MDP is defined as $\{\mathcal{S}, \mathcal{A}, \mathcal{R}\}$.

\textbf{1) States}

States represent the agent's observations of the current wireless environment. At time frame $t$, the state is defined as the channel power gain vector:
\begin{equation}
\mathbf{s}_t = \mathbf{h}_t = [h_{1,t}, h_{2,t}, \ldots, h_{N,t}]^T \in \mathbb{R}^N,
\label{eq:state_en}
\end{equation}
where $h_{i,t}$ represents the channel power gain between the $i$-th  WD and the AP. This state directly determines the local computation rate $r_{L,i}^*(a)$ and the offloading transmission rate $r_{O,i}^*(a, \tau_i)$. The state captures the instantaneous wireless channel conditions that vary across time frames due to fading effects, providing essential environmental information for offloading decisions.

To effectively process the high-dimensional channel state and extract critical features, we employ a Quantum Neural Network combined with attention mechanism  as the policy network. The quantum layer encodes the classical channel state into quantum features through parameterized quantum circuits, leveraging quantum superposition and entanglement to enhance feature representation capabilities. Subsequently, the multi-head attention mechanism dynamically weighs these quantum features to capture the importance and correlations among different channel dimensions. This hybrid architecture enables the model to adaptively focus on the most informative channel conditions, thereby improving decision quality in time-varying wireless environments.

\textbf{2) Actions}

Actions represent the binary task offloading decisions for all devices. At time frame $t$, the action is defined as:
\begin{equation}
\mathbf{a}_t = \mathbf{x}_t = [x_{1,t}, x_{2,t}, \ldots, x_{N,t}]^T \in \{0,1\}^N,
\label{eq:action_en}
\end{equation}
where $x_{i,t} \in \{0,1\}$ indicates whether device $i$ offloads its task to the AP ($x_{i,t}=1$) or computes locally ($x_{i,t}=0$). The action space size is $|\mathcal{A}| = 2^N$, which grows exponentially with the number of devices.

To efficiently explore this large discrete action space, the policy network first outputs a continuous probability vector $\mathbf{p}_t \in [0,1]^N$, where $p_{i,t}$ represents the probability that device $i$ should offload its task. This continuous representation is then converted into $K$ discrete binary action candidates through the UGQ quantization method. This quantization strategy effectively balances exploitation of the learned policy with exploration of alternative decisions, enabling the agent to discover high-quality offloading strategies.

\textbf{3) Reward}

Reward quantifies the quality of the offloading decision under the current channel state. Once the offloading decision $\mathbf{x}_t$ is fixed, the original mixed-integer problem (P1) reduces to a convex optimization subproblem (P2) concerning the continuous resource allocation variables $\{a, \boldsymbol{\tau}\}$. This subproblem can be efficiently solved using bisection search based on Lagrangian duality. The immediate reward at time frame $t$ is defined as the optimal objective value of subproblem (P2) for the given offloading decision $\mathbf{x}_t$, representing the maximized weighted sum computation rate:
\begin{equation}
\mathcal{R}_t = \max_{\boldsymbol{\tau}, a} Q(\mathbf{h}_t, \mathbf{x}_t, \boldsymbol{\tau}, a),
\label{eq:reward_en}
\end{equation}
where $a^*$ and $\boldsymbol{\tau}^* = [\tau_1^*, \ldots, \tau_N^*]^T$ are the optimal time allocations obtained by solving subproblem (P2). This reward function directly measures the system performance achieved by the selected offloading decision under the current channel conditions.

In practice, the agent generates $K$ candidate actions $\{\mathbf{x}_t^{(1)}, \mathbf{x}_t^{(2)}, \ldots, \mathbf{x}_t^{(K)}\}$ from the continuous probability vector. For each candidate, the corresponding reward is computed by solving the resource allocation subproblem. The action with the highest reward is then selected and executed:
\begin{equation}
\mathbf{x}_t^* = \operatorname*{arg\,max}_{\mathbf{x} \in \{\mathbf{x}_t^{(1)}, \ldots, \mathbf{x}_t^{(K)}\}} \mathcal{R}(\mathbf{h}_t, \mathbf{x}).
\label{eq:action_selection_en}
\end{equation}

The experience tuple $(\mathbf{h}_t, \mathbf{x}_t^*)$ is stored in a replay buffer for policy learning. The policy network parameters $\boldsymbol{\theta}$ are updated by minimizing the binary cross-entropy loss between the network output and the optimal actions. Through this supervised learning process, the agent learns to maximize the expected cumulative reward by generating near-optimal offloading decisions that adapt to time-varying wireless channels.

\subsection{ DNN and Uncertainty-Guided Quantization Method Design}

In the original DROO framework, the  Actor Module employs a fully connected DNN as the policy network. Its input consists of the current channel state and queue state, and it outputs the computation offloading preference vector for each user. Although DNNs possess strong nonlinear fitting capabilities, their feedforward structure struggles to effectively capture the dynamic temporal dependencies in the system state, such as channel fading, the continuity of task arrivals, and the evolution trend of queue states. Therefore, we replace the DNN with an RNN to explicitly model the temporal correlations within state sequences, thereby enhancing the policy network's adaptability to dynamic environments and its decision-making accuracy.

The RNN module designed in this paper adopts a multi-layer  GRU structure, with an input dimension equal to the number of users \( N \) and an output dimension also equal to \( N \). The network body contains two GRU layers and uses Batch Normalization (BatchNorm) to stabilize the training process. Finally, a fully connected layer and a Sigmoid activation function output the offloading probability vector for each user. Unlike the traditional DNN, which makes decisions based solely on the current state, the RNN memorizes historical system information through its hidden state, enabling the integration of multi-frame observations along the time dimension to more accurately predict the current optimal offloading strategy. Experimental results show that under identical training configurations, DROO with RNN significantly outperforms the original DNN-based method in terms of convergence speed and final normalized computation rate. Furthermore, the temporal modeling capability of the RNN also grants it stronger stability and robustness when facing non-independent and identically distributed (non-i.i.d.) or correlated dynamic task arrivals.

Building upon the original Order-Preserving Quantization method \ ref {alg:ugq}, this paper proposes a novel quantization method based on uncertainty guidance (Uncertainty-Guided Quantization, UGQ). The core idea of this method is: for each offloading decision probability value output by the deep neural network, the closer it is to 0.5, the more uncertain the model is regarding that dimension's decision. Therefore, when generating multiple candidate offloading actions, priority should be given to adjusting these uncertain dimensions, thereby achieving more effective and diverse exploration in the action space. Specifically, each time the UGQ method generates K candidate actions, it first selects a baseline action . Then, it sequentially selects several dimensions whose probabilities are closest to 0.5, using their probability values as pivot points. By introducing small random noise to adjust the threshold, it generates diverse candidate actions. If a newly generated action duplicates an existing one, a specific bit is forcibly flipped to maintain action diversity. This method not only retains the high efficiency of the original order-preserving approach but also significantly enhances action exploration capability by emphasizing uncertain dimensions and incorporating noise perturbation, thereby accelerating model convergence and improving final performance.

\begin{algorithm}
\caption{\enskip Uncertainty-Guided Quantization (UGQ) Method}\label{alg:ugq}
\begin{algorithmic}
  \Require Neural network output vector $\mathbf{m} \in [0,1]^N$, number of candidate actions $k \geq 1$, noise scale $\sigma$
  \Ensure Binary offloading action set $\mathcal{M} = \{\mathbf{x}_1, \mathbf{x}_2, \dots, \mathbf{x}_k\}$
  \State $\mathcal{M} \gets \emptyset$
  \State $\mathbf{x}_1 \gets [\mathbb{I}(m_i > 0.5)]_{i=1}^N$
  \State $\mathcal{M} \gets \mathcal{M} \cup \{\mathbf{x}_1\}$
  \If{$k = 1$}
    \State \Return $\mathcal{M}$
  \EndIf
  \State $\mathbf{d} \gets [|m_i - 0.5|]_{i=1}^N$
  \State $\mathbf{idx} \gets \text{argsort}(\mathbf{d})$
  \For{$i \gets 1$ to $k-1$}
    \State $p \gets m_{\mathbf{idx}[i \bmod N]}$
    \State $\tau \gets 0.5 + (p - 0.5) \times 0.7 + \mathcal{U}(-\sigma, \sigma)$
    \State $\mathbf{x} \gets [\mathbb{I}(m_j > \tau)]_{j=1}^N$
    \If{$\exists \mathbf{x}' \in \mathcal{M} \text{ s.t. } \mathbf{x} = \mathbf{x}'$}
      \State $f \gets \mathbf{idx}[(i+1) \bmod N]$
      \State $x_f \gets 1 - x_f$
    \EndIf
    \State $\mathcal{M} \gets \mathcal{M} \cup \{\mathbf{x}\}$
  \EndFor
  \State \Return $\mathcal{M}$
\end{algorithmic}
\end{algorithm}

\subsection{Attention Mechanism in the QNN Framework}

The integration of quantum computing principles into machine learning has given rise to QNNs, which leverage quantum superposition and entanglement to enhance computational efficiency and representational capacity. In reinforcement learning, Quantum Reinforcement Learning (QRL) combines quantum theory with RL to address challenges such as the exploration–exploitation trade-off and the curse of dimensionality in large state–action spaces \cite{4579244}. A key component in QRL is the Variational Quantum Circuit (VQC), also referred to as a parameterized quantum circuit or QNN, which maps classical inputs into quantum states and extracts features through quantum measurements \cite{10821103}.In quantum computation, the basic unit of information is the quantum bit (qubit). Unlike a classical bit, a qubit can exist in a superposition state:
\begin{align}
|\psi\rangle = \alpha|0\rangle + \beta|1\rangle,
\end{align}

where \(\alpha, \beta \in \mathbb{C}\) satisfy \(|\alpha|^2 + |\beta|^2 = 1\). The probability of observing the qubit in state \(|0\rangle\) or \(|1\rangle\) is given by \(|\alpha|^2\) and \(|\beta|^2\), respectively. An \(n\)-qubit system can represent a superposition of \(2^n\) states:

\begin{align}
|\phi\rangle = \sum_{x=0}^{2^n-1} C_x |x\rangle,
\end{align}

where \(C_x\) are complex probability amplitudes.

Quantum gates perform unitary transformations on qubits. Commonly used parameterized gates in VQCs include rotation gates \(R_x(\theta), R_y(\theta), R_z(\theta)\) and entanglement gates such as CNOT. A typical VQC consists of three stages: (1) state encoding \(U_{\text{enc}}\), which embeds classical data into a quantum state; (2) variational layers \(U_{\text{var}}(\theta)\) with trainable parameters; and (3) measurement, which yields classical outputs through expectation values .

For a classical input \(\mathbf{x} \in \mathbb{R}^d\), the quantum feature map can be expressed as:

\begin{align}
\mathbf{q} = \langle \psi(\mathbf{x}) | Z_i | \psi(\mathbf{x}) \rangle_{i=1}^{n_q},
\end{align}

where \(Z_i\) is the Pauli-\(Z\) operator on the \(i\)-th qubit, and \(\mathbf{q} \in \mathbb{R}^{n_q}\) is the extracted quantum feature vector.

In QRL, the policy or value function can be approximated using a VQC. Recent studies have shown that QRL can achieve comparable or superior performance to classical RL with fewer parameters, thanks to quantum parallelism and entanglement \cite{4579244}. Moreover, attention mechanisms—originally developed in classical deep learning—have been integrated into quantum models to enhance feature selection and contextual modeling. For example, Quantum Self-Attention Networks (QSAN) and Quantum Kernel Self-Attention Networks (QKSAN) have demonstrated improved performance in tasks such as image classification and combinatorial optimization \cite{10821103}.

Inspired by recent advances in quantum–classical hybrid neural networks, we enhance our QNN framework by incorporating a multi‑head self‑attention module that operates on the classical side. This hybrid architecture is designed to combine the high‑dimensional representational power of parameterized quantum circuits with the context‑aware feature‑weighting capability of modern attention mechanisms, leading to more robust and expressive models for sequential decision‑making tasks.

The proposed QNN+Attention module comprises two principal components: an efficient quantum encoding layer and a classical multi‑head self‑attention block. The overall data‑flow is illustrated in Fig.\ref{fig:a}.

Quantum Encoding Layer:The input feature vector \(\mathbf{x} \in \mathbb{R}^d\) is first projected into a lower‑dimensional space suitable for quantum processing via a classical feed‑forward network:

\begin{align}
\mathbf{x}' = \text{FFN}(\mathbf{x}) = \text{ReLU}\bigl(\mathbf{W}_c^{(2)}\;\text{ReLU}\bigl(\mathbf{W}_c^{(1)}\mathbf{x} + \mathbf{b}_c^{(1)}\bigr) + \mathbf{b}_c^{(2)}\bigr) \in \mathbb{R}^{n_b},
\end{align}

where \(\mathbf{W}_c^{(1)}, \mathbf{W}_c^{(2)}\) and \(\mathbf{b}_c^{(1)}, \mathbf{b}_c^{(2)}\) are trainable weight matrices and biases, respectively. The output dimension \(n_b\) is chosen to match the number of qubits used in the subsequent quantum circuit.

The reduced vector \(\mathbf{x}'\) is then encoded into a quantum state \(|\psi(\mathbf{x}')\rangle\) through a parameterized quantum circuit (PQC). The PQC consists of the following layers:

\begin{enumerate}
    \item Input encoding layer: For each qubit $i$, two rotation gates are applied:
          
          where the input features are scaled by $\pi$ and $\pi/2$ to span the full rotation range.
    \item Entanglement layer: A linear chain of CNOT gates creates entanglement between adjacent qubits:
          \begin{align}
              \text{CNOT}(i, i+1), \quad i = 0, 1, \dots, n_b-2.
          \end{align}
    \item Variational layer: A set of trainable rotation gates $R_z(\theta_i)$ is applied to each qubit, where $\theta_i$ are learnable parameters $\boldsymbol{\theta}_Q \in \mathbb{R}^{n_b}$.
\end{enumerate}

The resulting quantum state is:
\begin{align}
|\psi(\mathbf{x}'; \boldsymbol{\theta}_Q)\rangle = U_{\text{var}}(\boldsymbol{\theta}_Q) \, U_{\text{ent}} \, U_{\text{enc}}(\mathbf{x}') \, |0\rangle^{\otimes n_b}.
\end{align}

Finally, a quantum feature vector \(\mathbf{q} \in \mathbb{R}^{n_q}\) (with \(n_q = n_b\)) is extracted by measuring the expectation values of Pauli‑\(Z\) operators on each qubit:
\begin{align}
\mathbf{q} = \Bigl[ \langle \psi(\mathbf{x}'; \boldsymbol{\theta}_Q) | Z_i | \psi(\mathbf{x}'; \boldsymbol{\theta}_Q) \rangle \Bigr]_{i=1}^{n_q},
\end{align}
where \(Z_i\) denotes the Pauli‑\(Z\) operator acting on the \(i\)‑th qubit. This expectation‑value vector captures the quantum‑mechanical properties of the encoded input and serves as a high‑dimensional feature representation for the subsequent classical attention block.

Multi‑Head Self‑Attention Block,the quantum feature vector \(\mathbf{q}\) is reshaped into a sequence of length 1  and passed through a multi‑head self‑attention module. Let the input sequence be denoted as \(\mathbf{Q}_{\text{in}} \in \mathbb{R}^{1 \times n_q}\) (or \(\mathbb{R}^{L \times n_q}\)). 

For each attention head \(h = 1, \dots, H\), learnable linear projections generate the query, key, and value matrices:
\begin{align}
\mathbf{Q}_h = \mathbf{Q}_{\text{in}} \mathbf{W}_h^Q, \quad
\mathbf{K}_h = \mathbf{Q}_{\text{in}} \mathbf{W}_h^K, \quad
\mathbf{V}_h = \mathbf{Q}_{\text{in}} \mathbf{W}_h^V,
\end{align}
where \(\mathbf{W}_h^Q, \mathbf{W}_h^K, \mathbf{W}_h^V \in \mathbb{R}^{n_q \times d_k}\) are trainable weight matrices, and \(d_k = n_q / H\) is the dimension per head (ensuring that the total number of parameters is kept manageable).

The attention weights for head \(h\) are computed via the scaled dot‑product attention:
\begin{align}
\mathbf{A}_h = \text{softmax}\!\left( \frac{\mathbf{Q}_h \mathbf{K}_h^\top}{\sqrt{d_k}} \right) \in \mathbb{R}^{1 \times 1} \;(\text{or } \mathbb{R}^{L \times L}).
\end{align}

The output of head \(h\) is then:
\begin{align}
\mathbf{Z}_h = \mathbf{A}_h \mathbf{V}_h \in \mathbb{R}^{1 \times d_k} \;(\text{or } \mathbb{R}^{L \times d_k}).
\end{align}

All heads are concatenated along the feature dimension and projected back to the original dimension \(n_q\):
\begin{align}
\mathbf{Z} = \bigl[ \mathbf{Z}_1 \| \mathbf{Z}_2 \| \cdots \| \mathbf{Z}_H \bigr] \mathbf{W}^O \in \mathbb{R}^{1 \times n_q},
\end{align}
where \(\mathbf{W}^O \in \mathbb{R}^{(H \cdot d_k) \times n_q}\) is another trainable weight matrix.

\begin{table*}[!t]%
\centering %
\caption{Comparison of Four Algorithms on 30 Devices\label{tab:comparison_algorithms}}%
\begin{tabular*}{\textwidth}{@{\extracolsep\fill}lllll@{\extracolsep\fill}}
\toprule
\textbf{Method} & \textbf{Average Normalized Rate}  & \textbf{Average Time per Channel (s)}  & \textbf{Total Time (s)}   \\
\midrule
RNN+UGQ & 0.998401 & 0.016182 & 647.272567 \\
DNN+OP  & 0.987103 & 0.017022 & 680.879886 \\
DNN+UGQ & 0.994818 & 0.019143 & 765.727181 \\
RNN+OP  & 0.996838 & 0.012821 & 512.845740 \\
\bottomrule
\end{tabular*}
\end{table*}


\begin{figure}[htbp]
\centering
\caption{Performance comparison with 10 devices}
\label{fig:combined10}
\subfloat[Comparison of Normalization Rates of four algorithms on 10 devices\label{fig:Normalization_Rates_10}]{
    \includegraphics[width=0.45\textwidth]{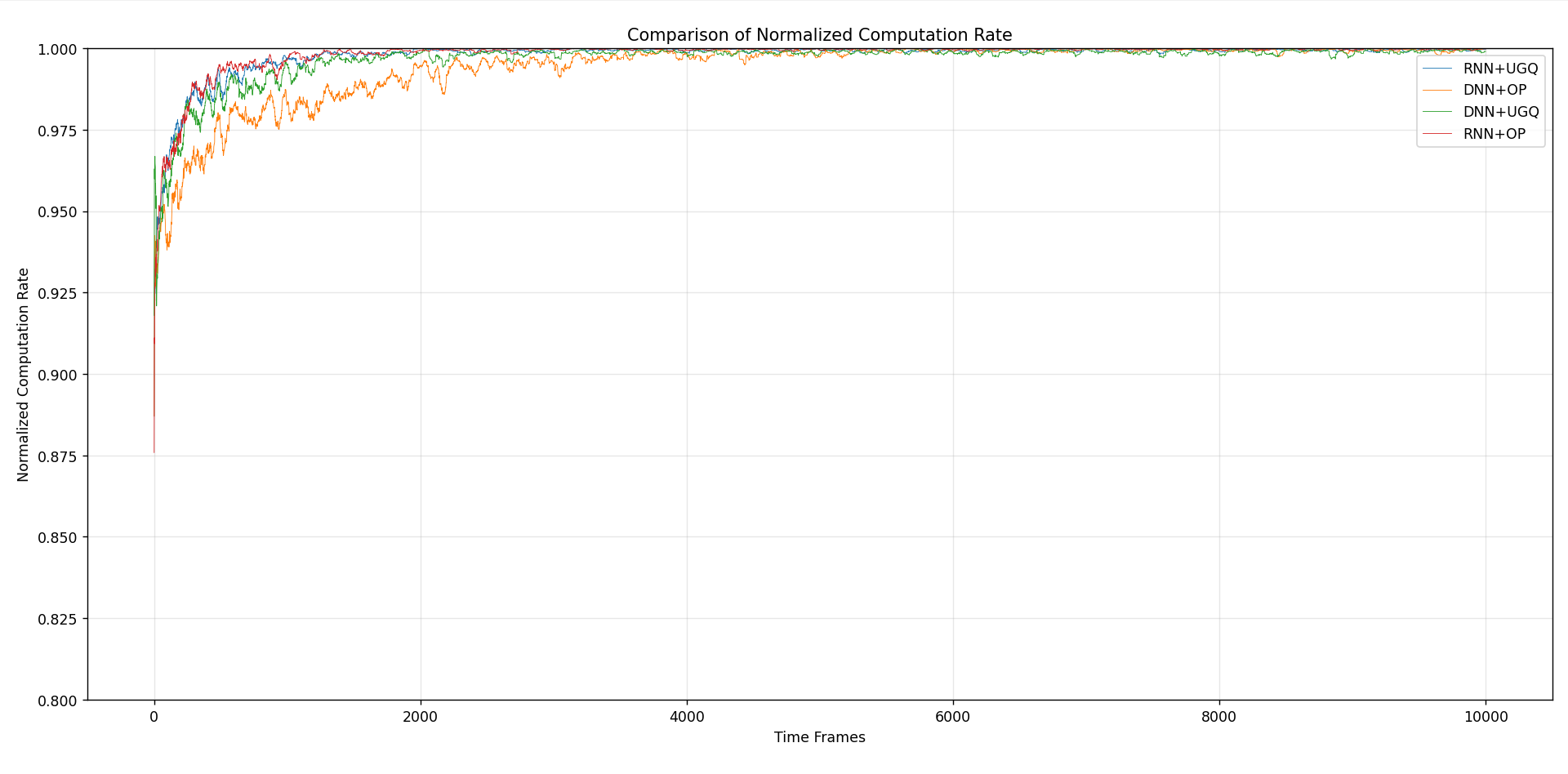}
}
\hfill
\subfloat[Comparison of loss functions of four algorithms on 10 devices\label{fig:loss_functions_10}]{
    \includegraphics[width=0.45\textwidth]{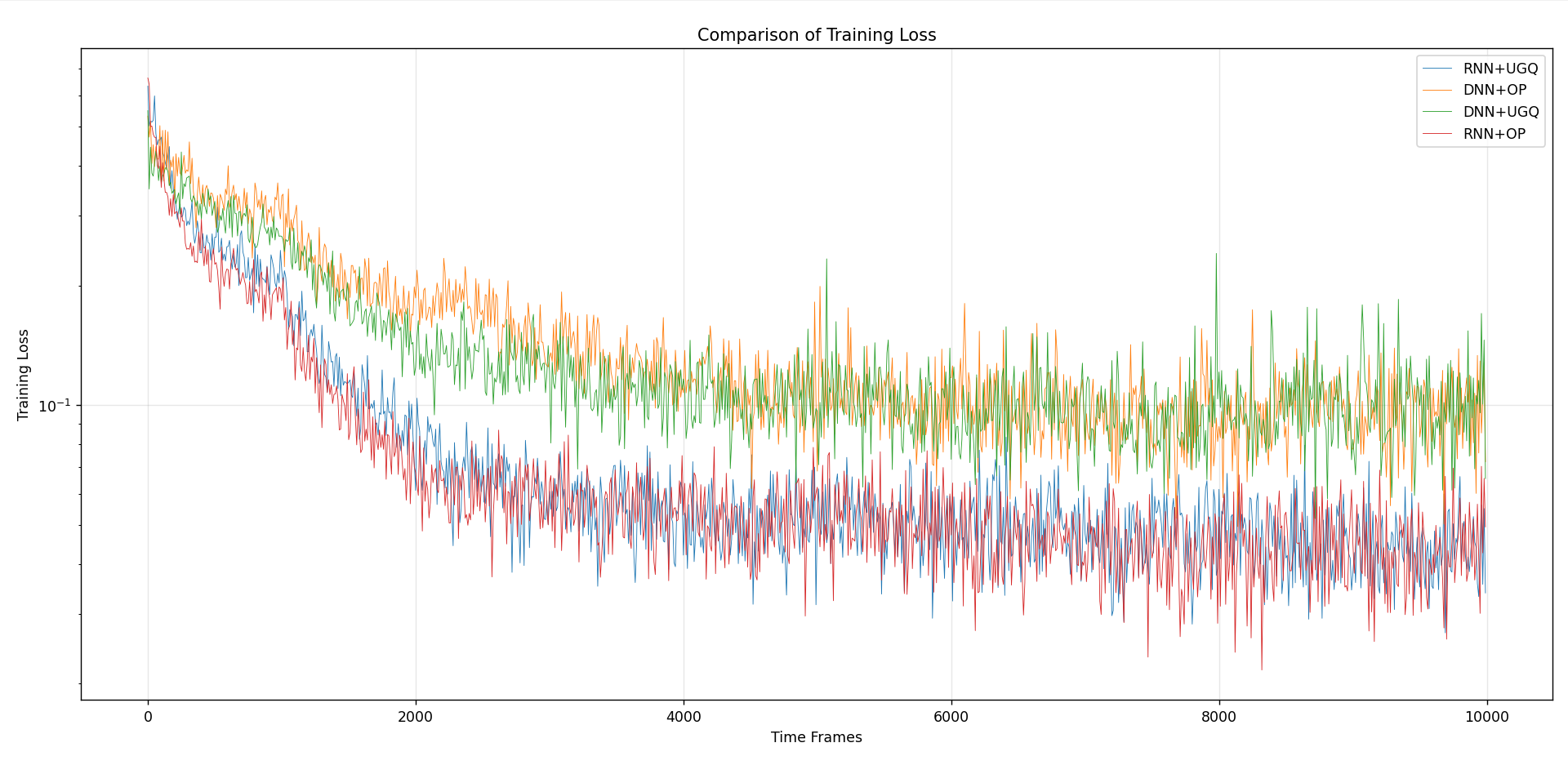}
}
\end{figure}

\section{Experiment and Results}\label{sec5}
\subsection{Experiments settings}
This section evaluates the performance of the proposed QAROO algorithm through simulation experiments. The experiments are conducted under the same wireless powered mobile edge computing system model as the original DROO algorithm to ensure fairness in comparison. Specific parameters are as follows: the energy-transmitting AP adopts the Powercast TX91501-3W with a transmit power \( P = 3\) W; each WD uses a P2110 energy harvester with an energy harvesting efficiency \( \mu = 0.51\). The distance \( d_i \) from the \( i \)-th WD to the AP is uniformly distributed within \((2.5, 5.2)\) meters. The average channel gain \( h_i \) is calculated based on the free-space path loss model, with an antenna gain \( A_d = 4.11\), carrier frequency \( f_c = 915\) MHz, and path loss exponent \( d_e = 2.8\). The time-varying channel gain is generated using the Rayleigh fading model, remaining constant within each time slot and independently varying across different time slots. The computational efficiency coefficient for all WDs is \( k_i = 10^{-26}\), the number of cycles required to process 1 bit of tasks is \( \phi = 100\), the bandwidth is \( B = 2\) MHz, the noise power is \( N_0 = 10^{-10}\), and user weights \( w_i \) are set to 1 (for odd-numbered users) and 1.5 (for even-numbered users) based on parity differences. The experiments are run on hardware equipped with a 13th Gen Intel Core i7-13700H processor and 16GB of memory. The software environment is built on Python 3.12.11 and PyTorch 2.7.1+cu128, with the quantum simulation part implemented using Qiskit.

At the algorithm level, the QAROO framework incorporates targeted designs from three aspects with corresponding parameters: the temporal modeling module employs a two-layer GRU network, each with 128 hidden units, and incorporates Dropout (0.1) and batch normalization to enhance generalization; the action generation module introduces an  UGQ strategy, starting with an initial threshold of 0.5, incorporating uniform noise perturbations of ±0.1, and forcing a flip in high-uncertainty dimensions when action repetition is detected to maintain exploration diversity; the feature extraction network adopts a quantum-classical hybrid attention architecture, where the quantum encoding layer uses an 8-qubit variational quantum circuit comprising input encoding, linear entanglement, and trainable rotation gates, while the classical attention module employs a 4-head self-attention mechanism with 2 dimensions per head, followed by two fully connected layers for decision output. During the training phase, the learning rate is set to 0.001, the experience replay buffer capacity is 1024, batch updates are performed every 10 time frames with a batch size of 128, and the Adam optimizer is used to optimize the binary cross-entropy loss.

\begin{figure}[htbp]
    \centering
    \caption{Performance comparison with 20 devices}
    \label{fig:combined20}
    \subfloat[Comparison of Normalization Rates of four algorithms on 20 devices\label{fig:Normalization_Rates_20}]{
        \includegraphics[width=0.45\textwidth]{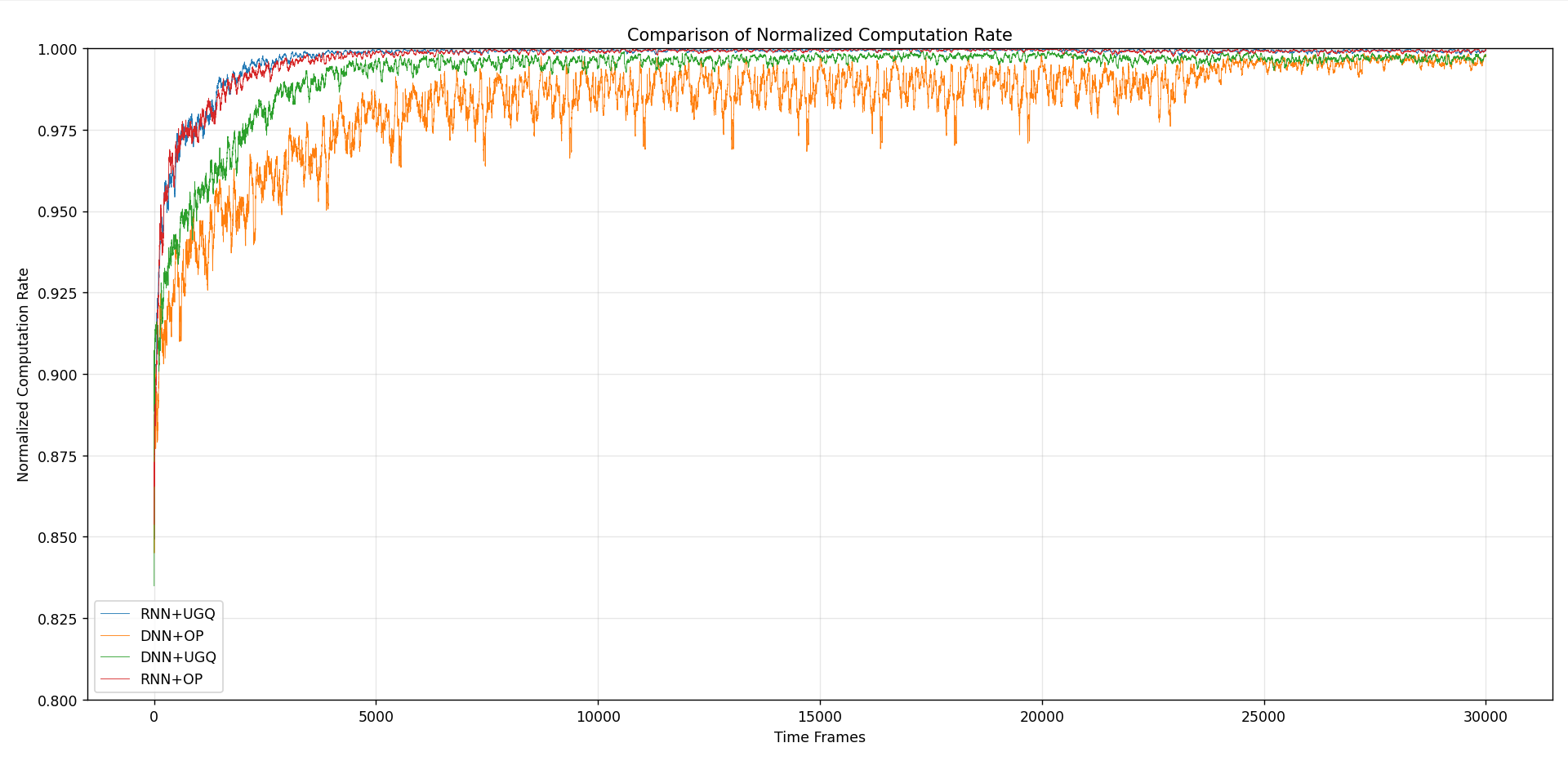}
    }
    \hfill
    \subfloat[Comparison of loss functions of four algorithms on 20 devices\label{fig:loss_functions_20}]{
        \includegraphics[width=0.45\textwidth]{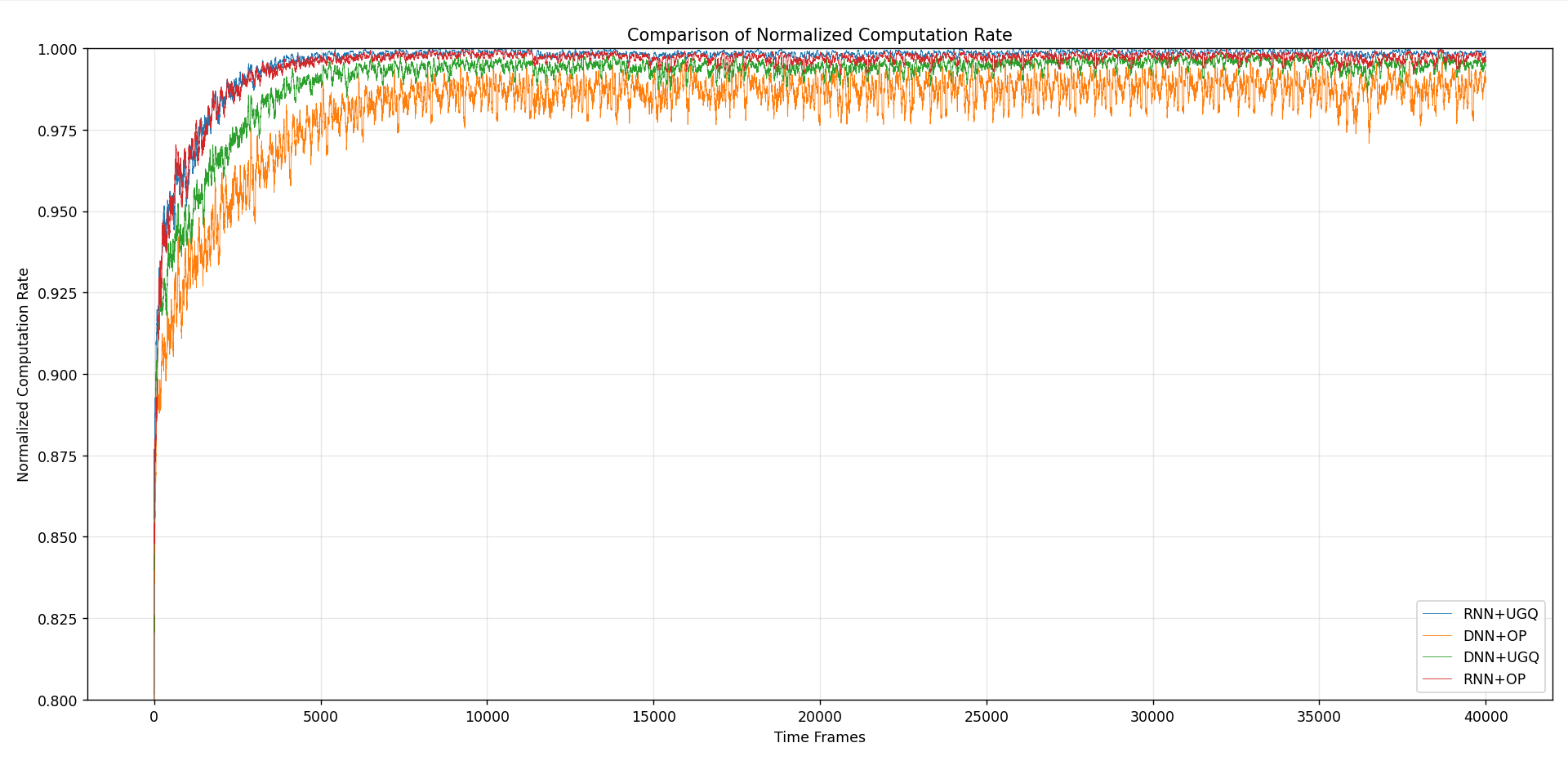}
    }
\end{figure}
\begin{figure}[htbp]
    \centering
    \caption{Performance comparison with 30 devices}
    \label{fig:combined30}
    \subfloat[Comparison of Normalization Rates of four algorithms on 30 devices\label{fig:Normalization_Rates_30}]{
        \includegraphics[width=0.45\textwidth]{picture/fig6.png}
    }
    \hfill
    \subfloat[Comparison of loss functions of four algorithms on 30 devices\label{fig:loss_functions_30}]{
        \includegraphics[width=0.45\textwidth]{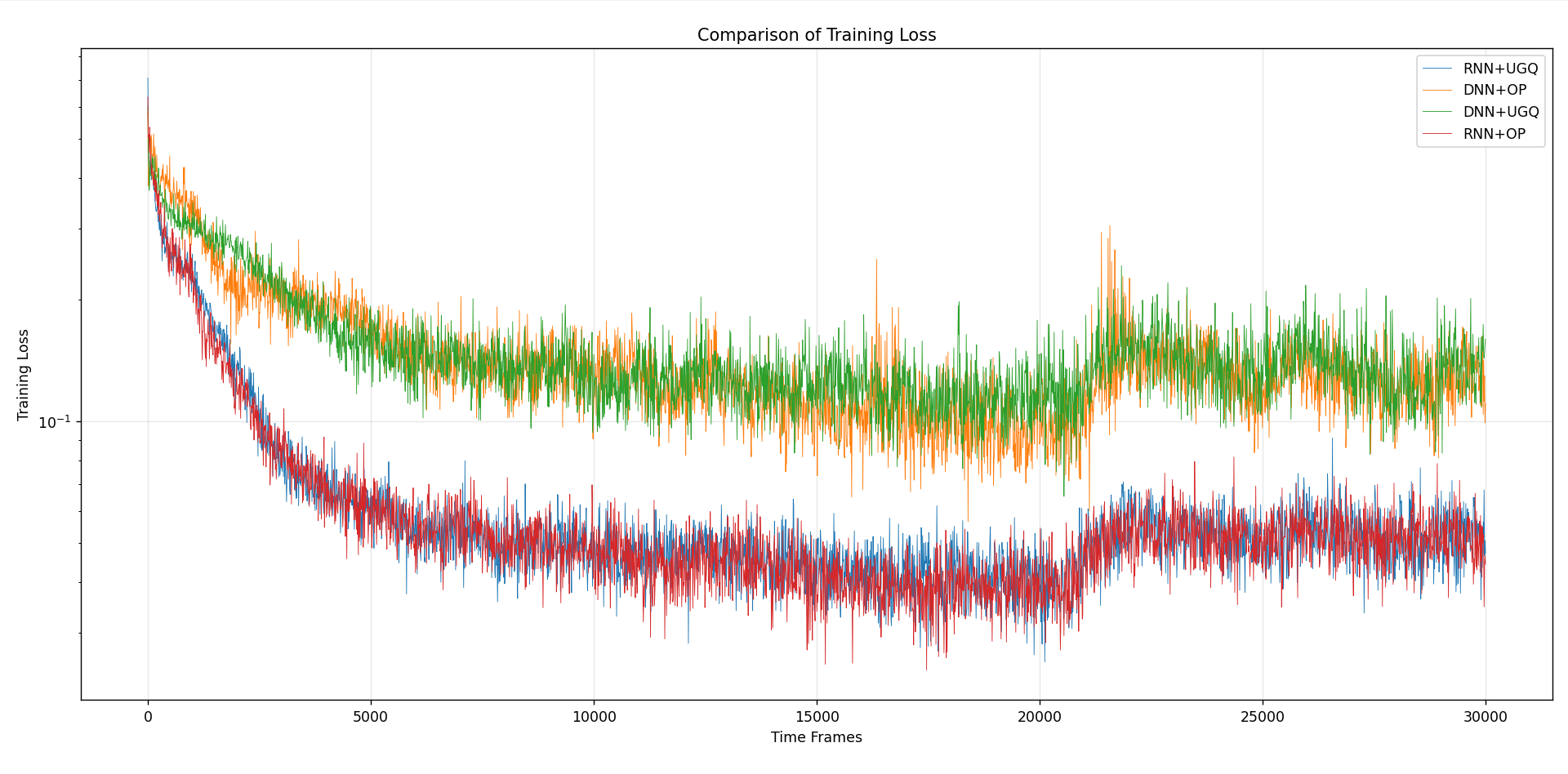}
    }
\end{figure}

As shown in Fig. \ref{fig:Normalization_Rates_30} performance comparison of normalization rates for four algorithms in the 30-device scenario and Table \ref{tab:comparison_algorithms} when the quantization method is fixed to either the traditional OP or the proposed UGQ the RNN-based framework demonstrates remarkable advantages over the DNN-based counterpart in terms of convergence dynamics and stability. Specifically during the early training phase for example in the first 1000 iterations the RNN-driven models exhibit a steeper upward trend in normalization rate achieving a near-optimal performance exceeding 0.99 for RNN+UGQ and RNN+OP much faster than the DNN-based algorithms DNN+OP reaches only 0.987103 after the same period. This accelerated convergence stems from the inherent temporal modeling capability of RNN equipped with a multi-layer GRU structure and Batch Normalization the RNN effectively captures the dynamic dependencies between consecutive channel states task arrival sequences and historical offloading decisions which the feedforward DNN structure fails to model. Unlike DNN that relies solely on instantaneous channel state information the RNN retains a hidden state to integrate multi-frame environmental observations enabling more accurate prediction of optimal offloading strategies in time-varying wireless fading channels.
Furthermore after convergence for example beyond 5000 iterations the RNN-based algorithms RNN+UGQ and RNN+OP exhibit significantly smaller fluctuations in normalization rate with a variance of less than 0.001 compared to DNN-based methods variance 0.003  for DNN+UGQ and 0.005  for DNN+OP reflecting superior robustness against non-stationary channel variations and correlated task arrivals. This stability is further validated by the loss function results in Fig. \ref {fig:loss_functions_30} the RNN experimental groups achieve a final loss value of 0.002  which is only half of the DNN groups 0.004  for DNN+UGQ and 0.006  for DNN+OP. The lower loss indicates that the RNN-based policy network generalizes better to unseen channel states reducing overfitting caused by the high-dimensional and dynamic nature of MEC offloading environments. As shown in Table \ref {tab:comparison_algorithms} RNN+UGQ achieves the highest average normalized rate 0.998401 among all four algorithms while RNN+OP maintains a competitive rate 0.996838 with the fastest average time per channel 0.012821 s confirming that RNN not only enhances performance but also preserves computational efficiency for real-time decision-making.

In the simulation environment with 10 devices as shown in Fig. \ref{fig:Normalization_Rates_10} the computational load of the system is relatively low due to the small number of connected devices. The action space size corresponding to 10 devices is only \(2^{10} = 1024\) which is a manageable scale for traditional quantization methods. For OP its fixed threshold-based quantization logic can effectively traverse the entire action space without obvious redundancy or missed exploration. During the offloading decision-making process both UGQ and OP can generate candidate actions that are sufficiently close to the optimal solution and the difference in the final normalization rate between the two methods is within a small range. Specifically for RNN-based models the normalization rate of both RNN+UGQ and RNN+OP reaches around 0.99 while for DNN-based models the normalization rate of both DNN+UGQ and DNN+OP is around 0.98. This negligible performance gap is mainly because the low computational load does not expose the inherent limitations of OP and the exploration capability of UGQ cannot be fully reflected in such a small-scale scenario.

However as the number of devices scales to 20 as shown in Fig. \ref{fig:Normalization_Rates_20} and 30 the system complexity and computational load increase sharply. The action space expands exponentially with the number of devices reaching approximately \(2^{20} = 1048576\) for 20 devices and \(2^{30} \approx 1073741824\) for 30 devices. At this point the limitations of OP become increasingly prominent. Its fixed threshold quantization method tends to generate homogeneous and redundant candidate actions which makes it difficult to fully explore the massive action space. A large number of potential high-quality offloading strategies are overlooked leading to insufficient optimization of the normalization rate. In contrast UGQ demonstrates significant advantages in large-scale scenarios by virtue of its unique uncertainty-guided mechanism. For 30 devices the improvement effect is more obvious the average normalized rate of RNN+UGQ reaches 0.998401 which is higher than 0.996838 of RNN+OP and the gap between DNN+UGQ and DNN+OP also expands to 0.007715 fully verifying that UGQ has stronger scalability and optimization capabilities in large-scale device deployment scenarios.

\begin{figure}[htbp]
    \centering
    \caption{Normalization rate of QAROO for different device quantities}
    \label{fig:QAROO}
    \subfloat[Comparison of Normalization Rates of four QAROO on 10 devices\label{fig:QAROO_10}]{
        \includegraphics[width=0.45\textwidth]{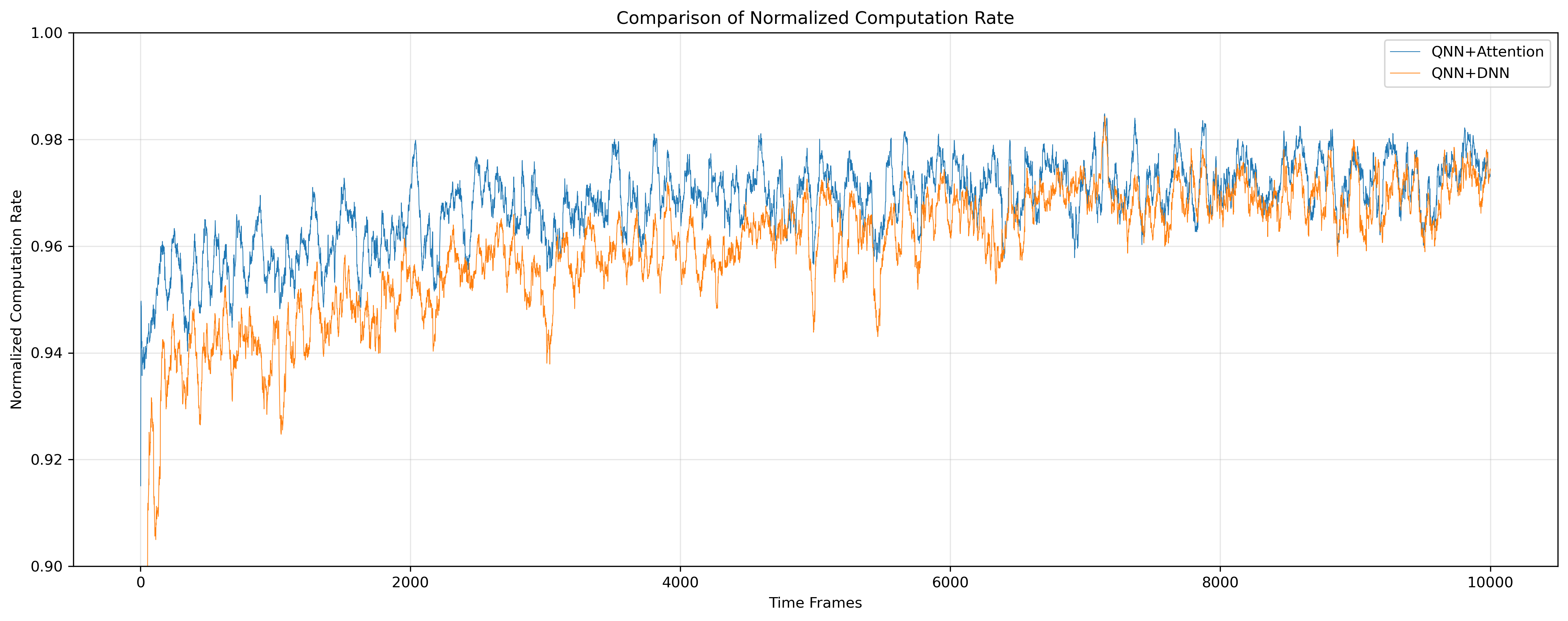}
    }
    \hfill
    \subfloat[Comparison of loss functions of four QAROO on 30 devices\label{fig:QAROO_30}]{
        \includegraphics[width=0.45\textwidth]{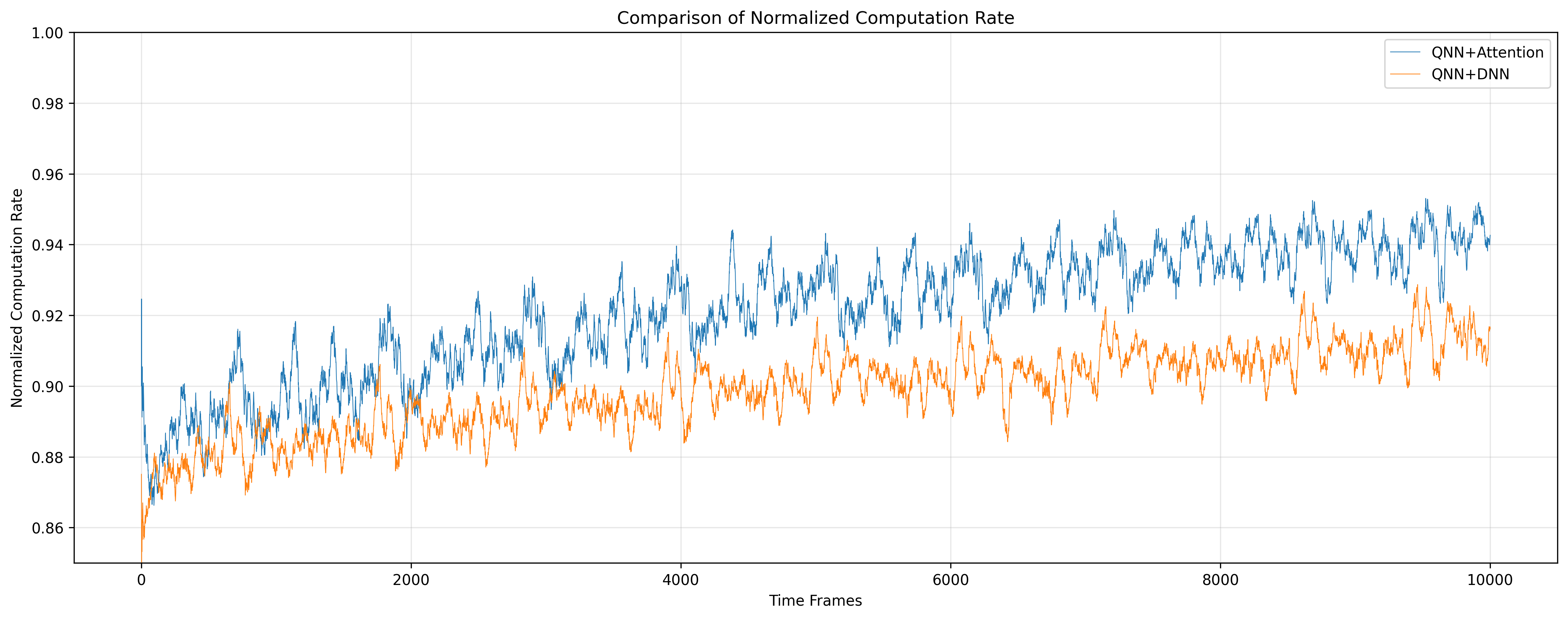}
    }
\end{figure}

In the 10-device scenario s, Fig. \ref{fig:QAROO_10}, both models use quantum superposition to map classical channel states into high-dimensional quantum features via quantum encoding layers. However, Quantum+DNN processes all quantum features uniformly through fully connected layers without distinguishing importance or inter-feature correlations, hindering rapid focus on critical cues . In contrast, Quantum+Attention integrates a multi-head self-attention block post-encoding, which projects quantum features into  matrices, computes dynamic attention weights, and prioritizes decision-relevant features. 

In the 30-device scenario , Fig. \ref{fig:QAROO_30}, Quantum+DNN’s limitations exacerbate: uniform processing of all quantum features causes information redundancy and inefficient optimization, leading to slower convergence and suboptimal performance. Quantum+Attention’s attention block excels in high-dimensional feature refinement by capturing complex inter-channel correlations , concentrating resources on informative features and suppressing redundancy.

\section{CONCLUSION}\label{sec6}
This paper proposes an online task offloading framework called QAROO for wireless-powered mobile edge computing networks. By integrating RNN, UGQ, and a hybrid QNN+Attention,the framework effectively addresses the coordinated optimization of computational and energy resources in dynamic channel environments. Specifically, the introduction of RNN enhances the system’s capability to model temporal channel states; the UGQ strategy improves action exploration diversity through dynamic threshold adjustment and stochastic perturbation; and the quantum-attention hybrid architecture significantly strengthens feature representation and decision robustness. Experiments demonstrate that RNN+UGQ outperforms traditional methods such as DROO in terms of normalized computation rate, convergence speed, and stability, particularly excelling in large-scale device scenarios. Future work will extend this framework to multi-cell MEC networks, partial task offloading, and mobility scenarios, and explore deployment on real quantum hardware with energy efficiency optimization.

\bibliographystyle{IEEEtran}
\bibliography{references}

\end{document}